\documentclass[runningheads]{llncs}

\usepackage{times}  % DO NOT CHANGE THIS
\usepackage{helvet} % DO NOT CHANGE THIS
\usepackage{courier}  % DO NOT CHANGE THIS
\usepackage[hyphens]{url}  % DO NOT CHANGE THIS
\usepackage{graphicx} % DO NOT CHANGE THIS
\usepackage{graphicx}  % DO NOT CHANGE THIS
\usepackage{times}
\usepackage{graphicx}     
\usepackage{amsmath}
\usepackage{txfonts}
\usepackage[ruled,vlined]{algorithm2e}

\DeclareMathAlphabet{\mathitbf}{OML}{cmm}{b}{it}

\newcommand{\sub}{_}
\def\su{^}
\newcommand{\actseq}{{\bar  a}}

\newcommand{\real}{{\mathbb{R}}}

\newcommand{\D}{{\cal D}}

\renewcommand{\L}{{\cal L}}

\newcommand{\N}{{\cal N}}

\renewcommand{\P}{{\cal P}}

\newcommand{\R}{{\cal R}}

\newcommand{\blemma}{\begin{lemma}}

\newcommand{\elemma}{\end{lemma}}

\newcommand{\bthm}{\begin{theorem}}
\newcommand{\ethm}{\end{theorem}}
\newcommand{\bprf}{\begin{proof}}
\newcommand{\eprf}{\end{proof}}
\newcommand{\bpro}{\begin{proposition}}
\newcommand{\epro}{\end{proposition}}
\newcommand{\bi}{\begin{itemize}}
\newcommand{\ei}{\end{itemize}}
\newcommand{\be}{\begin{enumerate}}
\newcommand{\ee}{\end{enumerate}}
\newcommand{\beq}{\begin{equation}}
\newcommand{\eeq}{\end{equation}}
\newcommand{\bcase}{\begin{cases}}
\newcommand{\ecase}{\end{cases}}
\renewcommand{\mathit}{\emph}

\begin{document}

%\begin{titlepage}       

%\setlength\titlebox{2.5in}       
	
%\title{Stochastic Uncertainty in the Situation Calculus}

% \title{Logic, Probability, Action: \\ A Situation Calculus and Cognitive Robotics  Perspective\thanks{The author was supported by a Royal Society University Research Fellowship.}}

\title{Logic, Probability and Action: \\ A Situation Calculus  Perspective\thanks{The author was supported by a Royal Society University Research Fellowship.}}

%\title{Knowledge, Probability and Actions: }

%\title{Incorporating Noise in }

%\title{Programs with Common Sense in Noisy Worlds for Cognitive Robotics}
%\title{Cognitive Robotics in Stochastic Domains}

\author{
Vaishak Belle\inst{}
}

\authorrunning{V. Belle}
\institute{University of Edinburgh, UK  \&  Alan Turing Institute, UK \\ 
\email{vaishak@ed.ac.uk}}

\maketitle
\begin{abstract} 
	
	The unification of logic and probability is a long-standing concern in AI, and more generally, in the philosophy of science. In essence,  logic provides an easy way to specify properties that must hold in every possible world, and probability allows us to further quantify the weight and ratio of the worlds that must satisfy a property.  
To that end, numerous developments have been undertaken, culminating in proposals such as probabilistic relational models. While this progress has been notable, a general-purpose first-order knowledge representation language to reason about probabilities and dynamics, including in continuous settings, is still to emerge. In this paper, we survey recent results pertaining to the integration of logic, probability and actions in the situation calculus, which is arguably one of the oldest and most well-known  formalisms. We then  explore    reduction theorems and programming interfaces for the language. These results are motivated in the context of cognitive robotics (as envisioned by Reiter and his colleagues) for the sake of concreteness. Overall, the advantage of proving results for such a general language is that it becomes possible to adapt them to any special-purpose fragment, including but not limited to popular probabilistic relational models.

\end{abstract}

% \category{I.2.4}{Artificial Intelligence}{Knowledge Representation Formalisms and Methods}
%
% \terms{Theory, Algorithms}
%
% \keywords{Cognitive robotics, high-level control, agent programming}
%
% \acmformat{Vaishak Belle, 2014. Pushing the Frontiers of High-level Robotic Control.}

\section{Introduction}

The unification of logic and probability is a long-standing concern in AI \cite{DBLP:journals/cacm/Russell15}, and more generally, in the philosophy of science \cite{demey2013logic}. The motivation stems from the observation  that (human and agent) knowledge is almost always incomplete. It is then not enough to say that some formula \( \phi \) is unknown. One
must also know which of $\phi$ or $\lnot\phi$ is the more likely, and by how
much. On the more pragmatic side, when reasoning about uncertain propositions and statements, it is beneficial to be able to leverage the underlying relational structure. Basically, logic provides an easy way to specify properties that must hold in every possible world, and probability allows us to further quantify the weight and  ratio of the worlds that must satisfy a property. For example, the sibling relation is symmetric in every possible world, whereas the influence of smoking among siblings can be considered a statistical property, perhaps only true in 80\% of the worlds. 

Another   argument increasingly made in favor of unifying logic and probability is that perhaps it would help us enable an apparatus analogous to Kahneman's so-called \textit{System 1}  versus \textit{System 2} processing in human cognition \cite{kahneman2011thinking}. That is, we want to interface  experiential and reactive processing (assumed to be handled by some data-driven probabilistic learning methodology) with cogitative processing (assumed to be handled by a deliberative reasoning methodology). 

% However, by unifying logic and probability we may also be able to make progress towards another question: what if the knowledge required for a task at hand is not obtained from observed examples?

To that end, numerous developments have been undertaken in AI. Closely following Bayesian networks \cite{pearl1988probabilistic,lerner2002monitoring}, and particle filters
\cite{dean1991planning,fox2003bayesian}, the areas of statistical relational learning and probabilistic relational modeling  \cite{de2011statistical,getoor2007introduction} emerged, and have been very successful. Since the world is rarely static, the application of such proposals to dynamic worlds has also seen many successes, e.g., \cite{Nitti:2016aa,tran2008event}. However, these closely follow propositional representations, such as Bayesian networks, using logic purely for templating purposes (i.e., syntactic sugar in programming language parlance). So, although the progress has been notable, a general-purpose first-order knowledge representation language to reason about probabilities and dynamics, including in continuous settings, is still to emerge. 

In the early days of the field, approaches such as
\cite{174658} provided a logical language that allowed one to reason about the probabilities of atoms, which could be further combined over logical connectives. That work has inspired numerous extensions for reasoning about dynamics. But this has been primarily in the propositional setting \cite{van2009dynamic,Halpern:1993:KPA:153724.153770,kushmerick1995algorithm}, or with discrete probabilistic models \cite{thielscher:AI01}. (See \cite{belle2018reasoning} for extended discussions.) 
In this paper, we survey recent results pertaining to the integration of logic, probability and actions in the situation calculus \cite{McCarthy:69,reiter2001knowledge}. The situation calculus is  one of the oldest and most well-known knowledge representation formalisms. In that regard, the results illustrate that we obtain perhaps the most expressive formalism for reasoning about degrees of belief in 
the presence of noisy sensing and acting. For that language, we then explore  reduction theorems and  programming interfaces. Of course, the advantage of proving results for such a general language is that it becomes possible to adapt them to any special-purpose fragment, including but not limited to popular probabilistic relational models. 

To make the discussion below concrete, we motivate one possible application of such a language: {\it cognitive robotics}, as envisioned by Reiter \cite{reiter2001knowledge} and further discussed in  \cite{lakemeyer2007chapter}. This is clearly not the only application of a language such as the situation calculus, which has found applications in areas such as service composition, databases, automated planning,  decision-theoretic reasoning and multi-agent systems  \cite{reiter2001knowledge,van2008handbook}. 

% Correspondingly, many of the reduction theorems considered for this language are applicable to any special-purpose fragment, which is seen to include popular probabilistic relational models.

\section{Motivation: Cognitive Robotics} % (fold)
\label{sec:motivation_cognitive_robotics}

% section motivation_cognitive_robotics (end)

The design and control of autonomous agents, such as robots, has been a major concern in artificial intelligence since the early days \cite{McCarthy:69}.  Robots can be viewed as systems that need to act purposefully in open-ended environments, and so are required to exhibit everyday commonsensical behavior. 
For the most part, however, traditional robotics has taken a ``bottom-up'' approach \cite{thrun2005probabilistic} focusing on low-level sensor-effector feedback.  Perhaps the most dominant reason for this is that controllers for  physical robots need to address the noise in effectors and sensors, often characterized by continuous probability distributions, which significantly complicates the reasoning and planning problems faced by a robot. 
% Incidentally, much of the advances in robotics benefits from methodologies developed in the AI communities, such as Bayesian approaches to handling probabilistic information.
While the simplicity of Bayesian statistics, defined over a fixed number of (propositional) random variables, has enabled the successful  handling of  probabilistic information in robotics modules,
% enables flexible implementations for a variety of sensorimotor interactions.
the flip side is that the applicability of contemporary methods is at the mercy of the roboticist's ingenuity. It is also unclear how precisely  commonsensical knowledge can be specified using conditional independences between  random variables while also accounting for how these dependencies further change as the result of actions. 

%(We will review notable exceptions in the penultimate section, and argue that these exceptions are, in fact, compatible with what is proposed in this work.) 

Cognitive robotics \cite{lakemeyer2007chapter}, as envisioned by Reiter and his colleagues \cite{reiter2001knowledge}, follows closely in the footsteps of McCarthy's seminal ideas \cite{McCarthy68programswith}: it 
takes the view that understanding the relationships between the beliefs of the agent and the actions at its disposal is key to a commonsensical robot that can operate purposefully   in uncertain, dynamic worlds. In particular, it considers the study of knowledge representation and reasoning problems faced by the  agent when attempting to answer questions such as \cite{levesque1998high}: \begin{quote}
\it 
\begin{itemize}
	\item to execute a program, what information does a robot need to have at the outset vs. the information that it can acquire en route by perceptual means?
	
	\item what does the robot need to know about its environment vs. what need only be known by the designer?
	
	\item when should a robot use perception to find out if something is true as opposed to reasoning about what it knows was true in the past?
\end{itemize}

 % That is, while traditional robotics attempts to leverage probabilistic reasoning, vision and  learning for stochastic control for a class of problems or domains, the cognitive element here is to relate these capabilities to the mental

\end{quote}
The goal, in other words, is to develop a theory of high-level control that maps the knowledge,  ignorance, intention and desires of the agent to appropriate actions. In this sense, cognitive robotics not only aims to connect to traditional robotics, which already  leverages probabilistic reasoning, vision and  learning for stochastic control, but also to relate to many other areas of AI, including automated planning, agent-oriented programming, {belief-desire-intention}  architectures, and formal epistemology. 

% \begin{itemize}
% 	\item for program execution, what information is needed from the designer versus the information that can be acquired by sensing?
%
% 	\item when should an agent reaffirm the truth value of a statement by sensing versus recovering the known value from memory?
%
% \end{itemize}

% Robots change the world by performing actions and learn about the world by sensing.
% In this sense, robotics research relates many themes in the field, such as automated planning, reasoning about incomplete information, probabilistic inference, vision and learning, among others.

In lieu of this agenda, many sophisticated control methodologies and formal accounts have emerged,  summarized in the following section. 
Unfortunately, despite the richness of these proposals, one criticism leveled at much of the work in cognitive robotics is that the theory is far removed from the kind of continuous uncertainty and noise seen in typical robotic applications. That is, the formal machinery of GOLOG to date does not address the complications due to noise and uncertainty in realistic robotic applications, at least in a way that relates these complications to what the robot believes, and how that changes over actions.   
% (There are notable exceptions \cite{boutilier2000decision,thielscher:ICAART10}, a discussion of which we defer to later.)
%but which still do not extend to the full logical language and arbitrary combinations of continuous fluents, effectors and sensors. See \cite{prego} for discussions.} 
The assumptions under which real-time behavior can be expected is also left open. For example, can standard probabilistic projection methodologies, such as Kalman and particle filters, be subsumed as part of a  general logical framework? 

%Much has to be worked out in that regard. 

%So, there are clearly at least two distinct ways of thinking of robotic control, but with 

% In this article, we report on a

The results discussed in this article can be viewed as a 
 research agenda that  attempts to bridge the gap between knowledge representation advances and robotic systems. By generalizing logic-based knowledge representation languages to reason about discrete and continuous probability distributions in the specification of both the initial beliefs of the agent and the noise in the sensors and effectors, the idea is  to contribute to commonsensical and provably correct high-level controllers for  agents in noisy worlds. 

% We are organized as follows. We will review the main strands of research in cognitive robotics in the following section. Then, we will discuss how some of these results have been revisited in a stochastic setting, and how, on the one hand, the generality is maintained, and on the other, tractable cases that connect to traditional robotics have been identified.
% We will then reflect on the core research questions that may  further close the gap between cognitive robotics and traditional robotics by reviewing some emerging trends in other subfields of AI.
% In the penultimate section, we will review related work and position cognitive robotics in the context of recent developments in other subfields of AI.
% Finally, we will conclude and end.
%

\section{Tools of the Trade} % (fold)
\label{sec:tools_of_the_trade}

% TODOs sit cal reiter version, FOL is enough Recap the main results that worked from cognitive robotics, 

To represent the beliefs and the actions, efforts in cognitive robotics would need to rely on a formal language of suitable expressiveness. Reiter's variant of the situation calculus has perhaps enjoyed the most success among first-order formalisms, although related proposals offer attractive properties of their own.\footnote{For example, the fluent calculus \cite{317154} offers an intuitive and simple state update mechanism in a first-order setting, and extensions of propositional dynamic logic \cite{721739} offer decidable formalisms.} Reiter's variant was also the language considered in a recent survey on  cognitive robotics \cite{lakemeyer2007chapter}, and so the reported results can easily be put into  context.\footnote{There has  been considerable debate on why a quantified relational language is crucial for knowledge representation and commonsense reasoning; see references in \cite{levesque2001logic,davis2015commonsense}, for example. Moreover, owing to the  generality of the underlying language, decidable variants can be developed (e.g., \cite{DBLP:journals/amai/GuS10,DBLP:journals/iandc/CalvaneseGMP18}). 
  % (e.g.,  \cite{DBLP:journals/amai/GuS10}).
}

  In this section, we will briefly recap some of the main foundational results discussed in \cite{lakemeyer2007chapter}. In a few cases, we report on recent developments expanding on those results. 

\newcommand{\ins}{S_{\!\textrm{0}}}

\subsection{Language} % (fold)
\label{sub:language}

% subsection language (end)

Intuitively, the language \( \L \) of the situation calculus \cite{McCarthy:69} is a
many-sorted dialect of predicate calculus, with sorts for \emph{actions},
\emph{situations} and \emph{objects} (for everything else, and includes the
set of reals \( \real \) as a subsort). A situation represents a world history
as a sequence of actions. A set of initial situations correspond to the ways
the world might be initially. Successor situations are the result of doing
actions, where the term \( do(a,s) \) denotes the unique situation obtained on
doing \( a \) in \( s. \) The term \( do(\actseq,s) \), where \( \actseq \) is
the sequence \( [a_1,\dots,a_n] \) abbreviates \(
do(a_n,do(\dots,do(a_1,s)\dots)). \) 
Initial situations are defined as those
without a predecessor, and we let the constant \( \ins \) denote the actual initial situation. See \cite{reiter2001knowledge} for a comprehensive treatment. 

% ,
% and here, we use the variable $\ivar$ to range over initial situations only.

The picture that emerges from the above is a set of trees, each rooted at an
initial situation and whose edges are actions. In general, we want the values
of predicates and functions to vary from situation to situation. For this
purpose, \( \L \) includes \emph{fluents} whose last argument is always a
situation. 

%Here we assume without loss of generality that all fluents arefunctional.

Following \cite{reiter2001knowledge}, dynamic domains in \(
\L \) are modeled by means of a \emph{basic action theory} \( \D \), which consists domain-independent foundational axioms, and a  domain-dependent  first-order initial theory \( \D \sub 0 \) (standing for what is true initially), and domain-dependent precondition and effect axioms, the latter taking the form of so-called successor state axioms that incorporates a monotonic solution to the frame problem \cite{reiter2001knowledge}. 

To represent knowledge, and how that changes, one appeals to the possible-worlds approach \cite{reasoning:about:knowledge}. 
% Following \cite{moore1985} and later \cite{citeulike:528170},
The idea is that there many different ways the world can be, where each world stands for a complete state of affairs. Some of these are considered possible by a putative agent, and they determine what the agent knows and does not know. Essentially, situations 
 can be viewed as possible worlds \cite{citeulike:528170}: a  special binary {fluent}  \( K \), taking two situation arguments  determines the accessibility relation between worlds. So, \( K(s',s) \) says that when the agent is at \( s \), he considers \( s' \) possible. \emph{Knowledge}, then, is simply truth at accessible worlds: \( \textit{Knows}(\phi,s) \doteq \forall s'.~ K(s',s) \supset \phi[s']. \) 
 
 Sensing axioms additionally capture the discovery of the truth values of fluents. For example, to check whether \( f \) is true at \( s, \) we would use: $\textit{SF}(\textit{sensetrue}_f,s) \equiv f(s) = 1.$ A successor state axiom formalizes the incorporation of these sensed values in the agent's mental state: \(
 		K(s', do(a,s)) \equiv\exists s''[K(s'', s) \land s' = do(a,s'')  ~\land   Poss(a, s'') ~\land (\textit{SF}(a,s'') \equiv \textit{SF}(a,s))]. 
 \) This says that if \( s'' \) is the predecessor of \( s', \) such that \( s'' \) was considered possible at \( s \), then   \( s' \) would be considered possible from \( do(a,s) \) contingent on sensing outcomes.

% \[
%  	 \begin{array}{l}
%  		K(s', do(a,s)) \equiv\exists s''[K(s'', s) \land s' = do(a,s'')  ~\land  \\ \hfill Poss(a, s'') ~\land (\textit{SF}(a,s'') \equiv \textit{SF}(a,s))].
%  	\end{array}
%  \]

\subsection{Reasoning Problems} % (fold)
\label{sub:reasoning_problems}

A fundamental problem underlying almost all applications involving basic action theories is \mathit{projection}. Given a sequence of actions $a\sub 1$ through $a\sub n$, denoted \( \actseq = [a\sub 1, \ldots, a\sub n], \) we are often interested in asking whether $\phi$ holds after these via entailment: \( \D \models \phi[do(\actseq, \ins)]? \) One of the main results by Reiter is the existence of a reduction operator called \mathit{regression} that eliminates the actions: \( \D \models \phi[do(\actseq, \ins) \textrm{ iff } \D\sub {\it una} \cup \D \sub 0 \models \R[\phi[do(\actseq,\ins)]. \) 
%Here, $\D \sub 0$ denotes the formulas that capture the world initially, $\D\sub {\it una}$ is a unique name axiom for actions to declare that actions instantiated from distinct function symbols cannot be the same, whereas $\D - (\D\sub {\it una} \cup \D \sub 0)$ includes the  foundational axioms, the successor state and precondition axioms. Thus, as far as projection is concerned, regression allows us to reason in a simple first-order, static setting, and ignore the dynamics and the second-order axioms for realizing the space of situations. 
Here, $\D\sub {\it una}$ is an axiom that declares that all named actions are unique, and  \( \R[\phi[do(\actseq,\ins)] \) mentions only a single situation term, \( \ins. \) 

In the worst case, regressed formulas are exponentially long in the length of the action sequence \cite{reiter2001knowledge}, and so it has been argued that for long-lived agents like robots, continually updating the current view of the state of the world, is perhaps better suited. Lin and Reiter \cite{lin1997progress} proposed a theory of progression that satisfies: 
\( \D \models \phi[do(\actseq, \ins) \textrm{ iff } \D\sub {\it una} \cup \P( \D \sub 0, \actseq) \models \phi[\ins]. \) Here \( \P( \D \sub 0, \actseq) \) is the updated initial theory that denotes the state of the world on doing \( \actseq. \) 
In general, progression requires second-order logic, but many special cases that are definable in first-order logic have since been identified (e.g., \cite{DBLP:conf/ijcai/LiuL09}). 

% subsection reasoning_problems (end)

\renewcommand{\mathit}{\textit}

\subsection{Closed vs Open Worlds} % (fold)
\label{sub:closed_vs_open_worlds}

 \( \D\sub 0 \) is assumed to be any set of first-order formulas, but then  computing its entailments, regardless of whether we appeal to regression or progression, would be undecidable. Thus, restricting the theory to be equivalent to a relational database is one possible tractable fragment, but this makes the closed world assumption which is not really desirable for robotics. A second possibility is to assume that at the time of query evaluation, the agent has complete knowledge about the predicates mentioned in the query. This leads to a notion of local completeness \cite{DBLP:conf/ijcai/GiacomoL99}. A third possibility is to provide some control over the computational power of the evaluation scheme, leading to a form of \mathit{limited reasoning}. First-order fragments such as {\it proper} and {\it proper}\( \su + \) \cite{DBLP:conf/kr/Levesque98,DBLP:conf/kr/Lakemeyer02}, which correspond to an infinite set of ground literals and clauses respectively, have been shown to work well with projection schemes for restricted classes of action theories  \cite{liu2005tractable,DBLP:conf/ijcai/LiuL09}. 
 
An altogether different and more general strategy for reasoning about incomplete knowledge is to utilize the epistemic situation calculus. A regression theorem was already proved in early work \cite{citeulike:528170}, and a progression theorem has been considered in  \cite{liuwen}. However, since propositional reasoning in epistemic logic is already intractable \cite{reasoning:about:knowledge}, results such as the \textit{representation theorem} \cite{levesque2001logic} that shows  how epistemic operators can be eliminated under \textit{epistemic closure} (i.e., knowing what one knows as well as what one does not know) needs to be leveraged at least. Alternatively, one could perhaps appeal to limited reasoning in the epistemic setting \cite{DBLP:conf/ecai/LakemeyerL12}.

% \footnote{The main concern with limited reasoning schemes is that on performing regression or progression, one needs to ensure that the regression formula or the progressed theory (respectively) is still in the fragment considered.
% % For example, it is shown in \cite{DBLP:journals/amai/GuS10} that  regression is also definable in a decidable situation calculus based on description logics.
% }

% subsection closed_vs_open_worlds (end)

\subsection{High-Level Control} % (fold)
\label{sub:high_level_control}

To program agents whose actions are interpreted over a basic action theory, \mathit{high-level programming languages} such as GOLOG emerged \cite{Levesque97-Golog}. These languages contained the usual programming features like sequence, conditional, iteration, recursive procedures, and concurrency but the key difference was that the primitive instruction was an action from a basic action theory. The execution of the program was then understood as \(
	\D\models Do(\delta, \ins, do(\actseq,\ins))
\) 
where \( \delta \) is a GOLOG program, and on starting from \( \ins, \) the program successfully terminates in \( do(\actseq,\ins) \). So, from \( \ins, \) executing the program leads to performing  actions \( \actseq \). 

As argued in \cite{lakemeyer2007chapter}, GOLOG programs can range from a fully deterministic instruction \( a\sub 1; \ldots; a\sub n \) to a general search \( \textbf{while } \neg \phi \textbf{ do } \pi a.~a \): the former instructs the agent to perform action \( a\sub 1 \),  then \( a\sub 2 \), and so on until \( a\sub n \) in sequence, and the latter instructs to try every possible action (sequence) until the goal is satisfied. It is between these two extremes where GOLOG is most powerful: it enables a partial specification of programs that can perform guided search for sub-goals in the presence of other loopy or conditional plans. 

To guide search in the presence of nondeterminism, rewards can be stipulated  on situations leading to a decision-theoretic machinery \cite{boutilier2000decision}. Alternatively, if the nondeterminism is a result of not knowing the true state of the world, sensing actions can be incorporated during program execution, leading to an \mathit{online} semantics for GOLOG  execution \cite{sardina2004semantics}. 

%The former amounts to knowing the plan so its uninteresting, and the latter 

% subsection high_level_control (end)

% TODO proper+ is PAC learnable 
% TODO commonsense not solved, ernie 
% TODO not clear we simply want to replicate traditional robotics, want to go beyond 

% section tools_of_the_trade (end)

\section{Tools Revisited} % (fold)
\label{sec:tools_revisited}

% TODO all the tools have been updated, first introduce language, open vs closed disappears (although credal comes up), regression progression (with special cases), golog with high-level control, 

% TODO maybe program synthesis / plan synthesis in core research questions 

In this section, we revisit the results from the previous section and discuss how these have been generalized to account for realistic,  continuous, models of noise.

Perhaps the most general formalism for dealing with {\it degrees of
  belief} in formulas, and in particular, with how degrees of belief should
evolve in the presence of noisy sensing and acting is the account proposed by
Bacchus, Halpern, and Levesque \cite{Bacchus1999171}, henceforth
BHL. Among its many properties, the BHL model shows precisely how beliefs can
be made less certain by acting with noisy effectors, but made more certain by
sensing (even when the sensors themselves are noisy). Not only  is it embedded in the rich theory of the situation calculus, including the use of Reiter's successor state axioms, it is also a  stochastic extension to the categorical epistemic situation calculus. The main advantage of a logical account like BHL is that it allows a
specification of belief that can be partial or incomplete, in keeping with
whatever information is available about the application domain.  It does not
require specifying a prior distribution over some random variables from which
posterior distributions are then calculated, as in Kalman filters, for example \cite{thrun2005probabilistic}.  Nor does it require specifying the conditional
independences among random variables and how these dependencies change as the
result of actions, as in the temporal extensions to Bayesian networks
\cite{pearl1988probabilistic}.  In the BHL model, some logical constraints are
imposed on the initial state of belief. These constraints may be compatible
with one or very many initial distributions and sets of independence
assumptions. (See \cite{belle2018reasoning} for extensive discussions.)  All the properties of belief will then follow at a corresponding
level of specificity. 

\subsection{Language} % (fold)
\label{sub:language}

The BHL model makes use of two distinguished binary fluents $p$
and $l$ \cite{bellelev13}. The \( p \) fluent determines a probability distribution on situations, by associating situations with \emph{weights}. More precisely, the term $p(s'\!,s)$ denotes the
relative \emph{weight} accorded to situation \( s' \) when the agent happens
to be in
situation \( s. \) Of course, \( p \) can be seen as a companion to \( K \). As one would for \( K, \)
the properties of $p$ in initial states, 
which vary from domain to domain, are specified with  axioms as part of \(
\D_0. \) The term $l(a,s)$ is intended to denote the likelihood of action $a$ in
situation $s$ to capture noisy sensors and effectors. For example, think of a sonar aimed at the wall, which gives a reading for the true value of a fluent \( f \) that corresponds to the distance between the robot and the wall.  Supposing the sonar's readings are subject to additive Gaussian noise. If now a reading of \( z \) were observed on the sonar, intuitively,  those situations where \( f = z \) should be considered more probable than those where  \( f \neq z \). Then we would have: \(
	l({\it sense}_f(z),s) = u 
	\equiv u = \N(z-f(s);0,1)). 
\)
Here, a standard normal is assumed, where the mean is 0, and the variance is 1.\footnote{If the specification of the $p$-axiom or the $l$-axiom includes disjunctions and existential quantifiers, we will then be dealing with uncertainty about distributions. See \cite{Belle:2015ae}, for example.
} Analogously, noisy effectors can be modeled using actions with double the  arguments: \(
	l(move(x,y),s) = u \equiv u = \N(y-x;0,1). 
\)
This says the difference between actual distance moved and the intended amount is normally distributed, corresponding to additive Gaussian noise. Such noise models can also be made context dependent (e.g., specifying the sensor's error profile to be worse for lower temperatures, where the temperature value is situation-dependent). In the case of noisy effectors, the successor state axioms have to be defined to use the second argument, as this is what actually happens at a situation \cite{belle2018reasoning}. 

\newcommand{\bel}{{\it Bel}}
Analogous to the notion of knowledge, the {\it degree of belief} in
$\phi$ in situation $s$ is defined as the weight of accessible worlds where \( \phi \) is true: \[
	\bel(\phi,s) \,\doteq\,	  \frac{1}{\gamma}\!\sum_{\{s':\phi[s'\!]\}} p(s'\!,s). 
\]
Here, \( \gamma \) is the normalization factor and corresponds to the numerator but with \( \phi \) replaced by \( {\it true}. \)
The change in \( p \) values over actions is specified using a successor state axiom, analogous to the one for \( K \): $p(s',do(a,s)) = u \,\,\equiv\exists s''~[s' = do(a,s'') \land {\it Poss}(a,s'') \land~ u = p(s'',s) \times l(a,s'')] \lor\,\, \neg\exists s''~[s' = do(a,s'') \land {\it Poss}(a,s'') 
           \land u = 0].$ 
This axioms  determines how \( l \) affects the \( p \)-value of successor situations.

% \begin{equation*}\label{eq:p ssa}
% 	\begin{array}{l}
% 		p(s',do(a,s)) = u \,\,\equiv\exists s''~[s' = do(a,s'') \land {\it Poss}(a,s'') \land~ u = p(s'',s) \times l(a,s'')] \\
% 		\qquad\qquad \qquad\qquad\lor\,\, \neg\exists s''~[s' = do(a,s'') \land {\it Poss}(a,s'')
%            \land u = 0].
% 	\end{array}
% \end{equation*}

As the BHL model is defined as a sum over possible worlds, it cannot actually handle Gaussians  and other  continuous distributions involving
  $\pi,$ $e,$ exponentiation, and so on. Therefore, BHL always consider
  discrete probability distributions that \emph{approximate} the continuous
  ones. However, this limitation was lifted in \cite{belle2018reasoning},  which shows how \( \bel \) is defined in continuous domains.

% subsection language (end)

\subsection{Reasoning Problems} % (fold)
\label{sub:reasoning_problems}

The projection problem in this setting is geared for reasoning about formulas that now mention \( \bel. \) In particular, we might be interested in knowing whether \( \D \models \bel(\phi,do(\actseq,\ins)) \geq r \)  for a real number \( r. \)

One reductive approach would be to translate both \( \D \) and \( \phi \), which would mention \( \bel, \) into a predicate logic formula. 
This approach, however, presents a serious computational problem because belief formulas expand into a large number of sentences, resulting in an enormous search space with initial and successor situations. The other issue with this approach is that sums (and integrals in the continuous case) reduce to complicated second-order formulas.

In \cite{BelleLevreg}, it is shown how Reiter's regression operator can be generalized to operate directly on \( \bel \)-terms. This involves appealing to the likelihood axioms. For example, imagine a robot that is uncertain about its distance   \( d \) to the wall, and the prior is a uniform distribution on the interval \( [2,12]. \) Assume the robot (noise-free)  moves away by 2 units and is now interested in the belief about \( d \leq 5. \) Regression would tell the robot that this is equivalent to its initial beliefs about \( d \leq 3 \) which here would lead to a value of \( .1 \). Imagine then the robot is also equipped with a sonar unit with additive Gaussian noise. After moving away by 2 units, if the sonar were now to provide a reading of  8, then regression would derive that belief about \( d \leq 5 \) is equivalent to \(
  {1}/{\gamma} \times  \int_2^3 .1 \times \N(6 - x;0,1) ~dx.
\) 
Essentially, the posterior belief about \( d \leq 5 \) is reformulated as the product of the prior belief about \( d \leq 3 \) and the likelihood of \( d \leq 3 \) given an observation of 6. 
%(This defines the upper bound on the integral; for the lower bound, note that for $d\leq 2$, the density is 0.) 
That is, observing 8 after moving away by 2 units  is equated here to observing 6 initially. (Here, \( \gamma \) is the normalization factor.)

% \[
%   \frac{1}{\gamma}  \int_2^3 .1 \times \N(6 - x;0,1).
% \]

%(While the initial result was limited to noisy sensing only, later work \cite{blprego,regression-and-progression-in-stochastic-domains} handles regression for beliefs over noisy effectors and sensors.) 

Progression too could potentially be addressed by expanding formulas involving \( \bel \)-terms, but it is far from clear what precisely this would look like. In particular, given  initial beliefs about fluents (such as the one about \( d \) earlier), we intuit that a progression account would inform us how this distribution changed. For example, on moving away from the wall by 2 units, we would now expect \( d \) to be uniformly distributed on the interval \( [4,14]. \) However, this leads to a complication: because if the robot had instead moved towards the wall by 4 units, then those points where \( d\in [2,4] \) initially are mapped to a single point \( d=0 \) that should then obtain a probability mass of .2, while the other points retain their initial density of .1. In \cite{blprog}, it is shown that for a certain class of basic action theories called \mathit{invertible  theories},  such complications are avoidable, and moreover, the  progressed database can be specified by means of simple syntactic manipulations. 

%These accounts of regression and progression can also be visualized using density change plots seen in Figure \ref{fig:density}, taken from \cite{blprog}. With regression, a query about the area under the blue curve in the left would reduce to a query about the area under the magenta rectangle. With progression, a query on doing a noisy move against the solid magenta is equivalent to testing the query directly against the blue curve in the right.
 
 %, which is the progressed database. 

% \begin{figure}[tbp]
% 			\centering
% 			\includegraphics[height=1.3in]{sense.pdf}
% 		\includegraphics[height=1.3in]{noisymove.pdf}
% 			\caption{\small Density change after repeated sensing (left) and a noisy move (right). Left: initially a uniform distribution (solid magenta), after sensing one (red circles), and after sensing twice (blue squares). Right: initially a normally distributed fluent (solid magenta) and after a noisy move  (blue squares).}
% 			\label{fig:density}
% 		\end{figure}

\subsection{Closed vs Open Worlds} % (fold)
\label{sub:open_vs_closed_worlds}

The closed vs open world discussion does not seem immediately interesting here, because, after all, the language is clearly open in the sense of not knowing the values of fluents, and according a distribution to these values. However, consider that the closed-world assumption was also motivated previously by computational concerns. In that regard, the above regression and progression results already studied special cases involving conjugate distributions \cite{box1973bayesian}, such as Gaussians which admit attractive analytical simplifications. For example, efficient Kalman filters \cite{thrun2005probabilistic} often make the assumption that the initial prior and the noise models are Gaussians, in which case the posterior would also be a Gaussian. In \cite{blprego}, it is further shown that when the initial belief is a Bayesian network, by way of regression, projection can be handled effectively by sampling. (That is, once the formula is regressed, the network is sampled and the samples are evaluated against the regressed formula.) 

In the context of probabilistic specifications, the notion of ``open"-ness can perhaps be interpreted differently. We can take this to mean that we do not know the distribution of the random variables, or even that the set of random variables is not known in advance. As argued earlier, this is precisely the motivation for the BHL scheme, and a recent modal reformulation of BHL illustrates the properties of such a language in detail \cite{Belle2017ac}. A detailed demonstration of how such specifications would work in the context of \mathit{robot localization} was given in \cite{Belle:2015ae}. 

The question of how to effectively compute beliefs in such rich settings is not clear, however. We remark that  various static frameworks have emerged for handling imprecision or uncertainty in probabilistic specifications \cite{DBLP:conf/ijcai/MilchMRSOK05,Cozman2000199,learning-tractable-credal-networks}. For example, when we have finitely many random variables but there is uncertainty about the underlying distribution, credal representations are of interest \cite{Cozman2000199}, and under certain conditions, they can be learned and reasoned with in an efficient manner \cite{learning-tractable-credal-networks}. On the other hand, when we have infinitely many random variables (but with a single underlying distribution), proposal such as \cite{DBLP:journals/cacm/Russell15} and \cite{belle2017weighted} are of interest, the latter being  a weighted representation inspired by \mathit{proper}\( ^+ \) knowledge bases. Extending these to  allow uncertainty about the underlying distribution may also be possible. Despite being static, by means of regression or progression, perhaps such open knowledge bases can be exploited for cognitive robotics applications, but that remains to be seen.  

% subsection open_vs_closed_worlds (end)

\subsection{High-Level Control} % (fold)
\label{sub:high_level_control}

A high-level programming language that deals with noise has to reason about two kinds of complications. First, when a noisy physical or sensing action in the program is performed, we must condition the next instruction on how the belief has changed as a result of that action. Second, because sensing actions in the language are of the form \mathit{sense(z)} that expects an input \( z, \) an \mathit{offline} execution would simulate possible values for \( z \) whereas an \mathit{online} execution would expect an external source to provide \( z \) (e.g., reading off the value of a sonar). We also would not want the designer to be needlessly encumbered by the  error profiles of the various effectors and sensors, so she has to be encouraged to program around \mathit{sense-act} loops; that is, every action sequence should be accompanied with a suitable number of sensing readings so that the agent is ``{confident}" (i.e., the distribution of the fluent in question is \mathit{narrow}) before performing more actions. In \cite{Belle:2015ab}, such a desiderata was realized to yield a stochastic version of knowledge-based programming \cite{reiter2001knowledge}. Primitive instructions are \mathit{dummy} versions of noisy actions and sensors; e.g., \mathit{move(x,y)} is simply \mathit{move(x)} and \mathit{sonar(z)} is simply \mathit{sonar}. The idea then is that the modeler simply uses these dummy versions as she would with noise-free actions, but the execution semantics incorporates the change in belief. It is further shown that program execution can be realized by means of a \mathit{particle filtering} \cite{thrun2005probabilistic} strategy: weighted samples are drawn from the initial beliefs, which correspond to initial situations, and on performing actions, fluent values at these situations are updated by means of the successor state axioms. The degree of belief in \( \phi \) corresponds to summing up the weights of samples where \( \phi \) is true. 

Such an approach can be contrasted with notable probabilistic relational modelling proposals such as \cite{Nitti:2016aa}: the difference mainly pertains to three  sources of generality. First, a  language like the situation calculus allows knowledge bases to be arbitrary quantificational theories, and BHL further allows uncertainty about the distributions defined for these theories. Second, the situation calculus, and by extension, GOLOG and the paradigm in \cite{Belle:2015ab} allows us to reason about non-terminating and unbounded behavior \cite{DBLP:conf/kr/ClassenL08}. Third, since an explicit belief state is allowed, it becomes possible to provide a systematic and generic treatment for multiple agents \cite{DBLP:conf/kr/KellyP08,Belle:2014aa}. 

On the issue of tractable reasoning, an interesting observation is that because these programs require reasoning with an explicit belief state \cite{reasoning:about:knowledge}, one might wonder whether the programs can be ``compiled" to a \mathit{reactive plan}, possibly with loops, where the next action to be performed depends only on the sensing information received in the current state. This relates knowledge-based programming to \mathit{generalized planning} \cite{DBLP:conf/aaai/Levesque96,siddthesis}, and of course, the advantage is also that numerous strategies have been identified to synthesize such loopy, reactive plans. Such plans are also shown to be sufficient for goal achievability \cite{DBLP:journals/ai/LinD98};  however, knowledge-based programs are known to be exponentially more succinct than  loopy, reactive plans \cite{lang2015probabilistic}. 
In \cite{DBLP:conf/aips/HuG13}, a generic algorithmic framework was  proposed to synthesize such plans in noise-free environments. How the correctness of such plans should be generalized to noisy environments was considered in \cite{Belle:2016ab,belle2018plans}. The  algorithmic synthesis problem was then considered in \cite{a-correctness-result-for-synthesizing-plans}.

% TODOs cite lang 
% TODOs talk about genealized planning, motivate lin and levesque, hu & giamcom 

% subsection high_level_control (end)

% TODO commonsense argument Ernie Davies, but on the other natural durative actions handeled 

% When it comes to  beliefs, and in particular how that changes after acting and sensing, the projection problem might be understood as  \emph{calculating} the degrees of belief in \( \phi \) after \( \actseq \); that is, find a real number \( r \) such that
% \begin{equation*}\label{eq:belief projection}
% 	     \D \models \bel(\phi,do(\actseq,\ins)) = r.
% \end{equation*}

 % \cite{regression-and-progression-in-stochastic-domains}.

% subsection reasoning_problems (end)

% section tools_revisited (end)

\section{Related Work and Discussions} % (fold)
\label{sec:lineage_and_challenges}

There are  many threads of research in AI, automated planning and robotics that are close in spirit to what is reported here. 
%We report on a few of these below. 
For example, belief update via the incorporation of sensor information has been considered in  probabilistic formalisms such as Bayesian networks \cite{pearl1988probabilistic,lerner2002monitoring},  
 Kalman and particle filters \cite{thrun2005probabilistic}. But these have difficulties handling strict uncertainty. Moreover, since rich models of actions are rarely incorporated, shifting conditional dependencies and distributions are hard to address in a general way. While there are graphical formalisms with an account of actions, 
such as  \cite{darwiche1994action,DBLP:conf/kr/HajishirziA10}, 
they too have  difficulties handling strict uncertainty and
quantification.
%(i.e., non-probabilistic and/or involving logical connectives). 
To the best of our knowledge, no existing probabilistic formalism handles changes in state variables like those possible in the BHL scheme. 
Related to these are  
relational probabilistic models \cite{ng1992probabilistic,DBLP:conf/ijcai/MilchMRSOK05,domingos2006unifying,DBLP:conf/uai/ChoiAH10}. Although limited accounts for dynamic domains are common here   \cite{lang2012jmlr,Nitti:2015aab}, explicit actions are seldom addressed in a general way. 
We refer interested readers to discussions in \cite{belle2018reasoning}, where differences are also drawn to prior  developments in reasoning about actions, including 
stochastic but non-epistemic GOLOG dialects \cite{DBLP:journals/igpl/GrosskreutzL03}. 

Arguably, many of the linguistic restrictions of such frameworks is often motivated by computational considerations. So what is to be gained by a general approach? This question is especially significant when we take into account that numerous  ``hybrid'' approaches have emerged over the years that  provide a bridge between a high-level language and a low-level operation  \cite{323918,lemaignan2010oro}. Our sense is that while these and other approaches are noteworthy, and are   extended in a modular manner to keep things tractable and workable on an actual physical robot, it still leaves a lot at the mercy of the roboticist's ingenuity. For example, extending an image recognition  algorithm to reason about a structured world is indeed possible, but it is more likely than not that this ontology 
is also useful for a number of other components, such as the robot's grasping arm; moreover, changes to one must mean changes to all. 
 % only applies to the vision system of the robot, and may differ from the ontology of the world according to its grasping arm.
 Abstracting a complex behavior module of a robot is a painstaking effort: often the robot's  modules are written in different programming languages with varying levels of abstraction, and to reduce these interactions to \mathit{atoms} in the high-level language  would require considerable know-how of the system. Moreover, although a roboticist can  abstract probabilistic sensors in terms of high-level categorical ones, there is loss in detail, as it is not clear at the outset which aspect of the sensor data is being approximated and by how much. Thus, all of these ``bottom-up'' approaches ultimately challenge the claim that the underlying theory is a genuine characterization of the agent. 
 
  %Consequently, claims of correctness at a wholistic level are hard to establish. 

In service of that, the contributions reported in this work attempt to express all the (inner and outer) workings of a robot in a \emph{single mathematical language}: a mathematical language that can capture rich structure as well as natively reason about the probabilistic uncertainty plaguing a robot; a mathematical language that can reason with all available information, some of which may be probabilistic, and some categorical; a mathematical language that can reason about the physical world at different levels of abstraction, in terms of objects, atoms, and whatever else physicists determine best describes our view of the world. Undoubtedly, given this glaring expressiveness, the agenda will raise significant challenges for the applicability of the proposal in contemporary robots, but our view is that, it will also engender novel extensions to existing algorithms to cope with the expressiveness. Identifying tractable fragments, for example, will engender novel theoretical work. As already discussed, many proposals from the statistical relational learning community are very promising in this regard, and are making steady progress towards the overall ambition. (But as discussed, they  fall short in terms of being able to reason about  non-terminating behavior, arbitrary first-order quantification, among other things,  and so identifying richer fragments is a worthwhile direction.)
% although they  fall short in being able to reason about unbounded action sequences (e.g., non-terminating behavior) and arbitrary first-order quantification. Thus, trying to extend the expressivity in that sense could lead a richer but still feasible knowledge representation language for modeling probabilistic knowledge but also reasoning about probabilistic events in a more powerful fashion.
It is also worth remarking that  the tractability of reasoning (and planning) has been the primary focus of much of the research in knowledge representation. The broader question of  how to learn models has received lesser attention, and this is precisely where statistical relational learning and related paradigms will prove useful \cite{belleinf}. (It would be especially interesting to consider relational  learning with neural modules \cite{de2019neuro}.)  Indeed, in addition to approaches such as \cite{Choi:2011fk,DBLP:conf/ijcai/ShiraziA05},  there have been a number of advances recently on learning dynamic  representations (e.g., \cite{Nitti:2016aa}), which might provide fertile ground to lift such ideas for cognitive robotics. 
% The task of learning in relational probabilistic models, including in dynamic domains, has been considered, for example, in  \cite{nitti2016learning,speichertlearning}.
 Computability results for qualitative learning in dynamic epistemic logic has been studied in \cite{bolander2015learning}. Recently, proper$^+$ knowledge bases were shown to be polynomial-time  learnable for querying tasks   \cite{belle2019implicitly}. Ultimately, learning may provide a means to coherently arrive at action descriptions at different levels of granularity from data \cite{davis2015commonsense}. 
% Another direction that a general formal approach is particularly insightful is in the context of multi-agent systems.
% One of the key goals in knowledge representation \cite{DBLP:journals/ci/LevesqueB87} is to investigate languages that are both attractive from a representational viewpoint but also computationally.
In the long term, the science of  building a robot, which currently is  more of an art, can perhaps be approached systematically. More significantly, through the agenda of cognitive robotics, we might gain deep insights on how commonsense knowledge and actions interact for general-purpose, open-ended robots. In that regard, the integration of logic, probability and actions will play a key role.

\small  
% Bibliography
% \bibliographystyle{ACM-Reference-Format-Journals}
% \bibliography{/Users/vaishakbelle/Dropbox/Papers/main}
%\small      
%\clearpage 
\bibliographystyle{abbrv}
\bibliography{main} 

\begin{thebibliography}{10}

\bibitem{Bacchus1999171}
F.~Bacchus, J.~Y. Halpern, and H.~J. Levesque.
\newblock Reasoning about noisy sensors and effectors in the situation
  calculus.
\newblock {\em Artificial Intelligence}, 111(1--2):171 -- 208, 1999.

\bibitem{belle2017weighted}
V.~Belle.
\newblock Weighted model counting with function symbols.
\newblock In {\em UAI}, 2017.

\bibitem{belle2018plans}
V.~Belle.
\newblock On plans with loops and noise.
\newblock In {\em AAMAS}, 2018.

\bibitem{belleinf}
V.~Belle.
\newblock Symbolic logic meets machine learning: A brief survey in infinite
  domains, 2020.

\bibitem{belle2019implicitly}
V.~Belle and B.~Juba.
\newblock Implicitly learning to reason in first-order logic.
\newblock {\em NeurIPS}, 2019.

\bibitem{Belle:2014aa}
V.~Belle and G.~Lakemeyer.
\newblock Multiagent only knowing in dynamic systems.
\newblock {\em Journal of Artificial Intelligence Research}, 49, 2014.

\bibitem{Belle2017ac}
V.~Belle and G.~Lakemeyer.
\newblock Reasoning about probabilities in unbounded first-order dynamical
  domains.
\newblock In {\em IJCAI}, 2017.

\bibitem{Belle:2016ab}
V.~Belle and H.~Levesque.
\newblock Foundations for generalized planning in unbounded stochastic domains.
\newblock In {\em KR}, 2016.

\bibitem{bellelev13}
V.~Belle and H.~J. Levesque.
\newblock Reasoning about continuous uncertainty in the situation calculus.
\newblock In {\em Proc. IJCAI}, 2013.

\bibitem{BelleLevreg}
V.~Belle and H.~J. Levesque.
\newblock Reasoning about probabilities in dynamic systems using goal
  regression.
\newblock In {\em Proc. UAI}, 2013.

\bibitem{blprog}
V.~Belle and H.~J. Levesque.
\newblock How to progress beliefs in continuous domains.
\newblock In {\em KR}, 2014.

\bibitem{blprego}
V.~Belle and H.~J. Levesque.
\newblock {PREGO}: {An Action Language for Belief-Based Cognitive Robotics in
  Continuous Domains}.
\newblock In {\em Proc. AAAI}, 2014.

\bibitem{Belle:2015ab}
V.~Belle and H.~J. Levesque.
\newblock Allegro: Belief-based programming in stochastic dynamical domains.
\newblock In {\em IJCAI}, 2015.

\bibitem{Belle:2015ae}
V.~Belle and H.~J. Levesque.
\newblock A logical theory of localization.
\newblock In {\em Studia Logica}. 2015.

\bibitem{belle2018reasoning}
V.~Belle and H.~J. Levesque.
\newblock Reasoning about discrete and continuous noisy sensors and effectors
  in dynamical systems.
\newblock {\em Artificial Intelligence}, 262:189--221, 2018.

\bibitem{bolander2015learning}
T.~Bolander and N.~Gierasimczuk.
\newblock Learning actions models: Qualitative approach.
\newblock In {\em International Workshop on Logic, Rationality and
  Interaction}, pages 40--52. Springer, 2015.

\bibitem{boutilier2000decision}
C.~Boutilier, R.~Reiter, M.~Soutchanski, and S.~Thrun.
\newblock Decision-theoretic, high-level agent programming in the situation
  calculus.
\newblock In {\em Proc. AAAI}, pages 355--362, 2000.

\bibitem{box1973bayesian}
G.~E.~P. Box and G.~C. Tiao.
\newblock {\em Bayesian inference in statistical analysis}.
\newblock Addison-Wesley, 1973.

\bibitem{323918}
W.~Burgard, A.~B. Cremers, D.~Fox, D.~H{\"a}hnel, G.~Lakemeyer, D.~Schulz,
  W.~Steiner, and S.~Thrun.
\newblock Experiences with an interactive museum tour-guide robot.
\newblock {\em Artif. Intell.}, 114(1-2):3--55, 1999.

\bibitem{DBLP:journals/iandc/CalvaneseGMP18}
D.~Calvanese, G.~{De Giacomo}, M.~Montali, and F.~Patrizi.
\newblock First-order \emph{{\(\mu\)}}-calculus over generic transition systems
  and applications to the situation calculus.
\newblock {\em Inf. Comput.}, 259(3):328--347, 2018.

\bibitem{DBLP:conf/uai/ChoiAH10}
J.~Choi, E.~Amir, and D.~J. Hill.
\newblock Lifted inference for relational continuous models.
\newblock In {\em Proc. UAI}, pages 126--134, 2010.

\bibitem{Choi:2011fk}
J.~Choi, A.~Guzman-Rivera, and E.~Amir.
\newblock Lifted relational kalman filtering.
\newblock In {\em Proc. IJCAI}, pages 2092--2099, 2011.

\bibitem{DBLP:conf/kr/ClassenL08}
J.~Cla{\ss}en and G.~Lakemeyer.
\newblock A logic for non-terminating golog programs.
\newblock In {\em KR}, pages 589--599, 2008.

\bibitem{Cozman2000199}
F.~G. Cozman.
\newblock Credal networks.
\newblock {\em Artificial Intelligence}, 120(2):199 -- 233, 2000.

\bibitem{darwiche1994action}
A.~Darwiche and M.~Goldszmidt.
\newblock Action networks: A framework for reasoning about actions and change
  under uncertainty.
\newblock In {\em Proc. UAI}, pages 136--144, 1994.

\bibitem{davis2015commonsense}
E.~Davis and G.~Marcus.
\newblock Commonsense reasoning and commonsense knowledge in artificial
  intelligence.
\newblock {\em Commun. ACM}, 58(9):92--103, 2015.

\bibitem{DBLP:conf/ijcai/GiacomoL99}
G.~De~Giacomo and H.~J. Levesque.
\newblock Projection using regression and sensors.
\newblock In {\em IJCAI}, 1999.

\bibitem{de2011statistical}
L.~De~Raedt and K.~Kersting.
\newblock Statistical relational learning.
\newblock In {\em Encyclopedia of Machine Learning}, pages 916--924. Springer,
  2011.

\bibitem{de2019neuro}
L.~De~Raedt, R.~Manhaeve, S.~Dumancic, T.~Demeester, and A.~Kimmig.
\newblock Neuro-symbolic= neural+ logical+ probabilistic.
\newblock In {\em NeSy'19 @ IJCAI}, 2019.

\bibitem{dean1991planning}
T.~Dean and M.~Wellman.
\newblock {\em Planning and control}.
\newblock Morgan Kaufmann Publishers Inc., 1991.

\bibitem{demey2013logic}
L.~Demey, B.~Kooi, and J.~Sack.
\newblock Logic and probability.
\newblock 2013.

\bibitem{domingos2006unifying}
P.~Domingos, S.~Kok, H.~Poon, M.~Richardson, and P.~Singla.
\newblock Unifying logical and statistical {AI}.
\newblock In {\em Proc. AAAI}, pages 2--7, 2006.

\bibitem{174658}
R.~Fagin and J.~Y. Halpern.
\newblock Reasoning about knowledge and probability.
\newblock {\em J. ACM}, 41(2):340--367, 1994.

\bibitem{reasoning:about:knowledge}
R.~Fagin, J.~Y. Halpern, Y.~Moses, and M.~Y. Vardi.
\newblock {\em Reasoning About Knowledge}.
\newblock {MIT} Press, 1995.

\bibitem{fox2003bayesian}
D.~Fox, J.~Hightower, L.~Liao, D.~Schulz, and G.~Borriello.
\newblock Bayesian filtering for location estimation.
\newblock {\em Pervasive Computing, IEEE}, 2(3):24--33, 2003.

\bibitem{getoor2007introduction}
L.~Getoor and B.~Taskar.
\newblock Introduction to statistical relational learning (adaptive computation
  and machine learning).
\newblock 2007.

\bibitem{DBLP:journals/igpl/GrosskreutzL03}
H.~Grosskreutz and G.~Lakemeyer.
\newblock ccgolog -- a logical language dealing with continuous change.
\newblock {\em Logic Journal of the IGPL}, 11(2):179--221, 2003.

\bibitem{DBLP:journals/amai/GuS10}
Y.~Gu and M.~Soutchanski.
\newblock A description logic based situation calculus.
\newblock {\em Ann. Math. Artif. Intell.}, 58(1-2):3--83, 2010.

\bibitem{DBLP:conf/kr/HajishirziA10}
H.~Hajishirzi and E.~Amir.
\newblock Reasoning about deterministic actions with probabilistic prior and
  application to stochastic filtering.
\newblock In {\em Proc. KR}, 2010.

\bibitem{Halpern:1993:KPA:153724.153770}
J.~Y. Halpern and M.~R. Tuttle.
\newblock Knowledge, probability, and adversaries.
\newblock {\em J. ACM}, 40:917--960, 1993.

\bibitem{721739}
A.~Herzig, J.~Lang, D.~Longin, and T.~Polacsek.
\newblock A logic for planning under partial observability.
\newblock In {\em Proc. AAAI / IAAI}, pages 768--773, 2000.

\bibitem{DBLP:conf/aips/HuG13}
Y.~Hu and G.~{De Giacomo}.
\newblock A generic technique for synthesizing bounded finite-state
  controllers.
\newblock In {\em ICAPS}, 2013.

\bibitem{kahneman2011thinking}
D.~Kahneman.
\newblock {\em Thinking, fast and slow}.
\newblock Macmillan, 2011.

\bibitem{DBLP:conf/kr/KellyP08}
R.~F. Kelly and A.~R. Pearce.
\newblock Complex epistemic modalities in the situation calculus.
\newblock In {\em KR}, 2008.

\bibitem{kushmerick1995algorithm}
N.~Kushmerick, S.~Hanks, and D.~Weld.
\newblock An algorithm for probabilistic planning.
\newblock {\em Artificial Intelligence}, 76(1):239--286, 1995.

\bibitem{DBLP:conf/ecai/LakemeyerL12}
G.~Lakemeyer and Y.~Lesp{\'e}rance.
\newblock Efficient reasoning in multiagent epistemic logics.
\newblock In {\em Proc. ECAI}, pages 498--503, 2012.

\bibitem{DBLP:conf/kr/Lakemeyer02}
G.~Lakemeyer and H.~J. Levesque.
\newblock Evaluation-based reasoning with disjunctive information in
  first-order knowledge bases.
\newblock In {\em Proc. KR}, pages 73--81, 2002.

\bibitem{lakemeyer2007chapter}
G.~Lakemeyer and H.~J. Levesque.
\newblock {Cognitive robotics}.
\newblock In {\em Handbook of Knowledge Representation}, pages 869--886.
  Elsevier, 2007.

\bibitem{lang2015probabilistic}
J.~Lang and B.~Zanuttini.
\newblock Probabilistic knowledge-based programs.
\newblock In {\em Twenty-Fourth International Joint Conference on Artificial
  Intelligence}, 2015.

\bibitem{lang2012jmlr}
T.~Lang, M.~Toussaint, and K.~Kersting.
\newblock Exploration in relational domains for model{--}based reinforcement
  learning.
\newblock {\em Journal of Machine Learning Research (JMLR)},
  13(Dec):3691{\&}{\#}8722;3734, 2012.

\bibitem{lemaignan2010oro}
S.~Lemaignan, R.~Ros, L.~M{\"o}senlechner, R.~Alami, and M.~Beetz.
\newblock Oro, a knowledge management platform for cognitive architectures in
  robotics.
\newblock In {\em IROS}, 2010.

\bibitem{lerner2002monitoring}
U.~Lerner, B.~Moses, M.~Scott, S.~McIlraith, and D.~Koller.
\newblock Monitoring a complex physical system using a hybrid dynamic bayes
  net.
\newblock In {\em Proc. UAI}, pages 301--310, 2002.

\bibitem{levesque1998high}
H.~Levesque and R.~Reiter.
\newblock High-level robotic control: Beyond planning.
\newblock Position paper at AAAI Spring Symposium on Integrating Robotics
  Research, 1998.

\bibitem{Levesque97-Golog}
H.~Levesque, R.~Reiter, Y.~Lesp\'erance, F.~Lin, and R.~Scherl.
\newblock Golog: A logic programming language for dynamic domains.
\newblock {\em Journal of Logic Programming}, 31:59--84, 1997.

\bibitem{DBLP:conf/aaai/Levesque96}
H.~J. Levesque.
\newblock What is planning in the presence of sensing?
\newblock In {\em Proc. AAAI / IAAI}, pages 1139--1146, 1996.

\bibitem{DBLP:conf/kr/Levesque98}
H.~J. Levesque.
\newblock A completeness result for reasoning with incomplete first-order
  knowledge bases.
\newblock In {\em Proc. KR}, pages 14--23, 1998.

\bibitem{levesque2001logic}
H.~J. Levesque and G.~Lakemeyer.
\newblock {\em {The logic of knowledge bases}}.
\newblock The MIT Press, 2001.

\bibitem{learning-tractable-credal-networks}
A.~Levray and V.~Belle.
\newblock Learning tractable credal networks.
\newblock In {\em AKBC}, 2020.

\bibitem{DBLP:journals/ai/LinD98}
F.~Lin and H.~J. Levesque.
\newblock What robots can do: Robot programs and effective achievability.
\newblock {\em Artif. Intell.}, 101(1-2):201--226, 1998.

\bibitem{lin1997progress}
F.~Lin and R.~Reiter.
\newblock {How to progress a database}.
\newblock {\em Artificial Intelligence}, 92(1-2):131--167, 1997.

\bibitem{DBLP:conf/ijcai/LiuL09}
Y.~Liu and G.~Lakemeyer.
\newblock On first-order definability and computability of progression for
  local-effect actions and beyond.
\newblock In {\em Proc. IJCAI}, pages 860--866, 2009.

\bibitem{liu2005tractable}
Y.~Liu and H.~Levesque.
\newblock {Tractable reasoning with incomplete first-order knowledge in dynamic
  systems with context-dependent actions}.
\newblock In {\em Proc. IJCAI}, pages 522--527, 2005.

\bibitem{liuwen}
Y.~Liu and X.~Wen.
\newblock On the progression of knowledge in the situation calculus.
\newblock In {\em IJCAI}, 2011.

\bibitem{McCarthy68programswith}
J.~McCarthy.
\newblock Programs with common sense.
\newblock In {\em Semantic Information Processing}, pages 403--418. MIT Press,
  1968.

\bibitem{McCarthy:69}
J.~McCarthy and P.~J. Hayes.
\newblock Some philosophical problems from the standpoint of artificial
  intelligence.
\newblock In {\em Machine Intelligence}, pages 463--502, 1969.

\bibitem{DBLP:conf/ijcai/MilchMRSOK05}
B.~Milch, B.~Marthi, S.~J. Russell, D.~Sontag, D.~L. Ong, and A.~Kolobov.
\newblock {BLOG}: Probabilistic models with unknown objects.
\newblock In {\em Proc. IJCAI}, pages 1352--1359, 2005.

\bibitem{ng1992probabilistic}
R.~Ng and V.~Subrahmanian.
\newblock Probabilistic logic programming.
\newblock {\em Information and {C}omputation}, 101(2):150--201, 1992.

\bibitem{Nitti:2016aa}
D.~Nitti.
\newblock {\em Hybrid Probabilistic Logic Programming}.
\newblock PhD thesis, KU Leuven, 2016.

\bibitem{Nitti:2015aab}
D.~Nitti, V.~Belle, and L.~D. Raedt.
\newblock Planning in discrete and continuous markov decision processes by
  probabilistic programming.
\newblock In {\em ECML}, 2015.

\bibitem{pearl1988probabilistic}
J.~Pearl.
\newblock {\em Probabilistic reasoning in intelligent systems: networks of
  plausible inference}.
\newblock Morgan Kaufmann, 1988.

\bibitem{reiter2001knowledge}
R.~Reiter.
\newblock {\em {Knowledge in action: logical foundations for specifying and
  implementing dynamical systems}}.
\newblock {MIT} Press, 2001.

\bibitem{DBLP:journals/cacm/Russell15}
S.~J. Russell.
\newblock Unifying logic and probability.
\newblock {\em Commun. {ACM}}, 58(7):88--97, 2015.

\bibitem{sardina2004semantics}
S.~Sardina, G.~De~Giacomo, Y.~Lesp{\'e}rance, and H.~J. Levesque.
\newblock On the semantics of deliberation in indigolog---from theory to
  implementation.
\newblock {\em Annals of Mathematics and Artificial Intelligence},
  41(2-4):259--299, 2004.

\bibitem{citeulike:528170}
R.~B. Scherl and H.~J. Levesque.
\newblock Knowledge, action, and the frame problem.
\newblock {\em Artificial Intelligence}, 144(1-2):1--39, 2003.

\bibitem{DBLP:conf/ijcai/ShiraziA05}
A.~Shirazi and E.~Amir.
\newblock First-order logical filtering.
\newblock In {\em Proc. IJCAI}, pages 589--595, 2005.

\bibitem{siddthesis}
S.~Srivastava.
\newblock {\em Foundations and Applications of Generalized Planning}.
\newblock PhD thesis, Department of Computer Science, University of
  Massachusetts Amherst, 2010.

\bibitem{317154}
M.~Thielscher.
\newblock From situation calculus to fluent calculus: state update axioms as a
  solution to the inferential frame problem.
\newblock {\em Artificial Intelligence}, 111(1-2):277--299, 1999.

\bibitem{thielscher:AI01}
M.~Thielscher.
\newblock Planning with noisy actions (preliminary report).
\newblock In {\em Proc. Australian Joint Conference on Artificial
  Intelligence}, pages 27--45, 2001.

\bibitem{thrun2005probabilistic}
S.~Thrun, W.~Burgard, and D.~Fox.
\newblock {\em Probabilistic Robotics}.
\newblock {MIT Press}, 2005.

\bibitem{tran2008event}
S.~D. Tran and L.~S. Davis.
\newblock Event modeling and recognition using markov logic networks.
\newblock In {\em Proc. ECCV}, pages 610--623, 2008.

\bibitem{a-correctness-result-for-synthesizing-plans}
L.~Treszkai and V.~Belle.
\newblock A correctness result for synthesizing plans with loops in stochastic
  domains.
\newblock {\em International Journal of Approximate Reasoning}, 2020.

\bibitem{van2009dynamic}
J.~Van~Benthem, J.~Gerbrandy, and B.~Kooi.
\newblock Dynamic update with probabilities.
\newblock {\em Studia Logica}, 93(1):67--96, 2009.

\bibitem{van2008handbook}
F.~Van~Harmelen, V.~Lifschitz, and B.~Porter.
\newblock {\em Handbook of knowledge representation}.
\newblock Elsevier, 2008.

\end{thebibliography}
% History dates
%\received{February 2007}{March 2009}{June 2009}

% Electronic Appendix

\end{document}